\documentclass[sigconf]{acmart}
\usepackage{subcaption}
\usepackage{stfloats} 

\AtBeginDocument{%
  }

\setcopyright{none}
\copyrightyear{2026}
\acmYear{2026}
\acmDOI{XXXXXXX.XXXXXXX}
\acmConference[HRI '26]{ACM/IEEE International Conference on Human-Robot Interaction}{March 16--19, 2026}{Edinburgh, United Kingdom}

\renewcommand\footnotetextcopyrightpermission[1]{}


\begin{document}

\title[Breathe with Me]{Breathe with Me: Synchronizing Biosignals for User Embodiment in Robots}
\thanks{This paper has been accepted for publication at the ACM/IEEE International Conference on Human-Robot Interaction (HRI 2026). The final version will appear in the ACM Digital Library.}

\author{Iddo Yehoshua Wald}
\authornote{Both authors contributed equally to this research.}
\email{wald@uni-bremen.de}
\orcid{0000-0001-7998-0133}
\affiliation{%
  \institution{University of Bremen}
  \city{Bremen}
  \country{Germany}
}

\author{Amber Maimon}
\authornotemark[1]
\email{maimon.amber@gmail.com}
\orcid{0000-0002-4592-0711}
\affiliation{%
  \institution{University of Bremen}
  \city{Bremen}
  \country{Germany}
}
\affiliation{%
  \institution{University of Haifa}
  \city{Haifa}
  \country{Israel}
}

\author{Shiyao Zhang}
\email{shiyao@uni-bremen.de}
\orcid{0000-0001-5965-0428}
\affiliation{%
  \institution{University of Bremen}
  \city{Bremen}
  \country{Germany}
}

\author{Dennis Küster}
\email{kuester@uni-bremen.de}
\affiliation{%
  \department{Cognitive Systems Lab}
  \institution{University of Bremen}
  \city{Bremen}
  \country{Germany}
}

\author{Robert Porzel}
\email{porzel@tzi.de}
\affiliation{%
  \department{The Digital Media Lab}
  \institution{University of Bremen}
  \city{Bremen}
  \country{Germany}
}

\author{Tanja Schultz}
\email{tanja.schultz@uni-bremen.de}
\affiliation{%
  \department{Cognitive Systems Lab}
  \institution{University of Bremen}
  \city{Bremen}
  \country{Germany}
}

\author{Rainer Malaka}
\email{malaka@uni-bremen.de}
\orcid{0000-0001-6463-4828}
\affiliation{%
  \department{The Digital Media Lab}
  \institution{University of Bremen}
  \city{Bremen}
  \country{Germany}
}

\renewcommand{\shortauthors}{Wald and Maimon et al.}

\settopmatter{authorsperrow=4}

\begin{abstract}

Embodiment of users within robotic systems has been explored in human–robot interaction, most often in telepresence and teleoperation. In these applications, synchronized visuomotor feedback can evoke a sense of body ownership and agency, contributing to the experience of embodiment. We extend this work by employing embreathment, the representation of the user's own breath in real time, as a means for enhancing user embodiment experience in robots. In a within-subjects experiment, participants controlled a robotic arm, while its movements were either synchronized or non-synchronized with their own breath. Synchrony was shown to significantly increase body ownership, and was preferred by most participants. We propose the representation of physiological signals as a novel interoceptive pathway for human–robot interaction, and discuss implications for telepresence, prosthetics, collaboration with robots, and shared autonomy.

\end{abstract}

\begin{teaserfigure}
  \includegraphics[width=\textwidth]{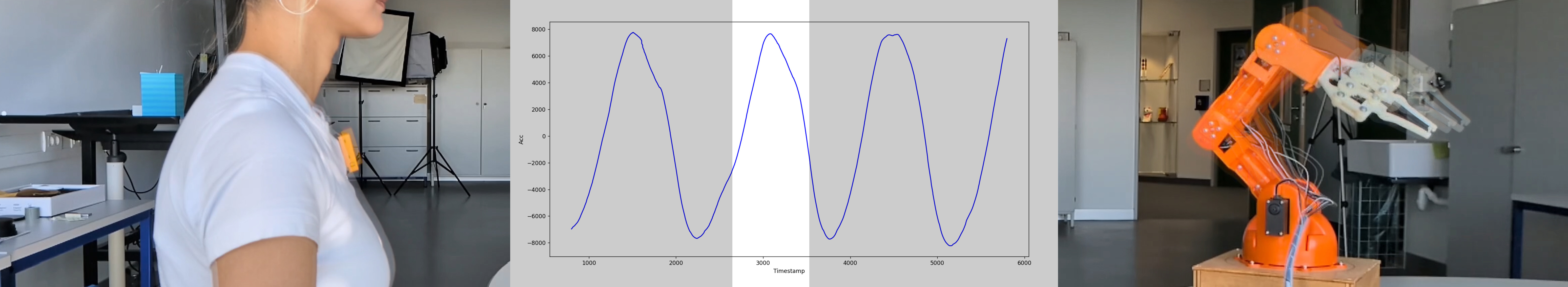}
  \caption{A user inhaling and exhaling with the robotic arms movement synced to its breath. In the middle is the signal from the respiration sensor placed on the user's chest.}
  \Description{A wide image is divided into three sections. On the left, a person in a white shirt is shown in profile, wearing a small respiration sensor on their chest. In the center, a line graph displays accelerometer data over time, showing a smooth, wave-like breathing pattern. On the right, a robotic arm is captured in multiple overlaid positions, illustrating synchronized movement that mimics the user's breathing cycle.}
  \label{fig:}
\end{teaserfigure}


\maketitle

\section{Introduction}

Embodiment of humans in robots is increasingly discussed as an evolving domain of human–robot interaction, specifically in the context of telepresence and teleoperation. In these settings, users control a remote robot or device while receiving synchronous sensory feedback. Numerous studies have shown that such coupling can elicit a striking sense of ownership and agency, whether toward humanoid robots~\cite{aymerich2015embodiment}, non-anthropomorphic robots~\cite{aymerich2017non}, virtual tools~\cite{jahanian_najafabadi_tool-use_2023} or avatars~\cite{aymerich2017object}. Functional congruence appears more critical than physical resemblance for the sense of embodiment, a point captured in the “functional body model hypothesis,” which explicitly argues that functionality drives embodiment more than visual similarity~\cite{aymerich2017non, aymerich2016role}. Participants often report that the teleoperated system feels like an extension of their own body, enhancing presence, fluency, and task performance~\cite{toet2020toward, sato2018body, nishio2012body, haans_embodiment_2012, falcone2024sense}. This body of work demonstrates that robotic embodiment is not only possible but can produce powerful benefits. Yet most approaches to date have relied on visuomotor synchrony, vibro-tactile feedback, or muscle-driven control, leaving other physiological pathways underexplored.

In this work, we utilize ``embreathment''~\cite{monti_embreathment_2020, wald_breathing_2023, wald_enriched_2025}—the synchronization of a user’s breathing with robotic behavior—as a novel pathway to support embodiment in active interaction. Breathing is a powerful interoceptive signal that continuously shapes bodily self-awareness and has been shown to modulate ownership illusions even in simple mannequin hand paradigms~\cite{kosuge2023respiratory}. Breathing-based interactions with robots have primarily been explored for affect regulation, such as reducing stress or supporting relaxation~\cite{matheus2022social, lotzin2025feasibility}, but their potential for enhancing user embodiment has not been tested. By coupling respiration to robotic motion in real time, we seek to leverage interoceptive synchrony to extend this technique beyond passive illusion paradigms and into active human–robot interaction.
A video demonstrating the system and study procedure is available online.\footnote{Video: https://osf.io/2eqcg/files/f6j9q}

\subsection{Research Questions}

This work addresses the following research questions:

\begin{itemize}
    \item \textbf{RQ1:} Can real-time breath synchronization between a user and a robot enhance the user's sense of embodiment in the robot?
    \item \textbf{RQ2:} How does breath synchronization influence broader perceptions of the robot, including self-efficacy and social attributes?
\end{itemize}

\subsection{Contributions}

This work makes the following contributions:

\begin{itemize}
    \item We introduce, the first to our knowledge, application of embreathment in robot operation, demonstrating that breath synchronization can enhance embodiment in human-robot interaction.
    \item We provide empirical evidence through a controlled experiment, showing that breath synchronization significantly increases body ownership of a robotic arm, with medium to large effect sizes on standardized embodiment measures.
    \item We discuss future applications of this approach, suggesting interoceptive-exteroceptive synchrony more generally and beyond breathing as a means for achieving user embodiment in HRI, as well as the potential of user embodiment in HRI for various applications.
\end{itemize}

\section{Related work}

\subsection{Sense of Embodiment}

The sense of embodiment is defined as being comprised of three main factors: (1) body ownership, (2) agency, and (3) self-location~\cite{kilteni_sense_2012}. Its study has roots in philosophy and cognitive science~\cite{Merleau_Ponty_2013, varela_embodied_2017}, and became an empirical focus in psychology and neuroscience through body ownership illusions such as the rubber hand illusion~\cite{botvinick1998rubber}. Subsequent work extended the concept into virtual reality, human–computer interaction, and prosthetics, where embodiment is tied to both functional and psychosocial outcomes~\cite{slater2010first, zbinden2022prosthetic, graczyk2018home, kocur_rubber_2022}. Embodiment in HRI is mostly discussed when referring to robots as embodied agents, which is the core differentiating feature of robots from other digital agents. In this work, we focus on the human side of embodiment, and the embodiment experience of a user in a robot. 

Embodiment is thought to be facilitated by expansion of the body schema. The body schema is a multisensory, action-oriented model of the body that the brain uses to predict and control movement, and when external cues align with this model, external objects such as tools or prosthetic limbs can be incorporated into it. This has been demonstrated behaviorally~\cite{cardinali2009tool, maravita_tools_2004, maravita_body_2006, jahanian_najafabadi_tool-use_2023} and neurally in premotor representations of peripersonal space~\cite{graziano1998spatial, graziano_attention_2015, graziano_attention_2017}.

At the bodily level, when incoming cues match the brain’s multisensory predictions, the system can claim an external object as part of the body. This has been shown through body ownership illusions over the years. The classic rubber hand illusion demonstrates how synchronous stimulation of a seen object and one’s hidden body elicits a sense of ownership and shifts perceived body position toward the external object, while asynchronous input abolishes the effect~\cite{botvinick1998rubber}. Comparable ownership effects have also been demonstrated in virtual reality settings, validating the transfer of these principles from the physical to the virtual domain~\cite{kocur_rubber_2022}. Beyond limb-level phenomena, full-body ownership illusions have been demonstrated, for example when participants experience a virtual or mannequin body from a first-person perspective as their own~\cite{ehrsson2007experimental, slater2010first}, indicating that embodiment principles generalize across scales of bodily representation. 

Ownership increases with temporal synchrony and plausible body geometry or kinematics, and drops with asynchrony or implausible configurations~\cite{makin2008other}. When the external body moves, matching one's voluntary actions to those movements elicits both ownership and agency. However, passive movement (where the external body moves without the person's intention) or anatomical incongruence can eliminate one component while preserving the other~\cite{kalckert_moving_2012}. This dissociation reveals that ownership and agency are distinct psychological processes that can operate independently.

Crucially, these findings demonstrate that embodiment depends more on functional integration, i.e., how well the external body responds to and extends one's intentions, than on physical resemblance. This supports the 'functional body model hypothesis,' which holds that functional congruence is ultimately more important than anatomical similarity for eliciting a sense of embodiment~\cite{aymerich2017non, aymerich2016role}.

\subsection{Physiological Interactions in HRI}

Physiological interactions refer to those that utilize processes arising from internal bodily functions, many of which are regulated by the autonomic nervous system. These are often not directly observable from the outside but can be sensed interoceptively. Common examples include heartbeat and respiration. Physiology in HRI has primarily been explored in three directions: (1) physiological computing, where body signals inform or adapt interaction; (2) incorporating physiological patterns into robots, for changing how they are perceived. e.g make them appear more lifelike to improve how humans perceive the robots; and (3) mimicking or following human signals in order to help users regulate these signals and associated states, such as breathing exercises and anxiety regulation.

Physiological computing represents a promising frontier in human-robot interaction, where psycho-physiological signals such as galvanic skin response, heart rate variability, and EEG are used to estimate human states and adapt robot behavior accordingly~\cite{savur2023survey}. By continuously monitoring biosignals during human–robot collaboration, these systems can infer attention, stress, workload, or trust levels and adjust robot motion or interaction style in real time, ultimately improving safety, comfort, and task fluency. Recent HRI datasets that combine physiological measurements with affect labels provide empirical support for this approach~\cite{heinisch2024physiological}. This growing body of evidence has strengthened arguments for closing the loop between physiological sensing and dynamic robot adaptation~\cite{kothig2021connecting}, pointing toward more responsive and human-centered robotic systems.

Shifting from sensing physiological signals to representing them, another use of physiology in HRI has been for affecting human perception of robots. This relates to how movement patterns that resemble biological motion can enhance perceived animacy~\cite{castro-gonzalez_effects_2016}. For example, adding breathing cues to collaborative robots has been shown to enhance measures of robot perception, such as intelligence, animacy, social presence, and likeability, highlighting how physiological rhythms can enrich interaction dynamics~\cite{terziouglu2020designing}. Breathing cues were also used for conveying emotional states, such as pleasure and arousal~\cite{klausen2022signalling}. Using participants’ own physiological signals, a pre-recorded vibrotactile pulse paced to each individual’s resting heart rate, but not live-synced, was found to enhance the robot’s perceived presence and lifelikeness, while also supporting emotion regulation~\cite{borgstedt2024soothing}.  

In relation to the ability to support emotion regulation, a third line of work mirrors human physiological signals to regulate user states such as anxiety, relaxation, or affect. Prior studies demonstrate that patterned cues can effectively reduce stress: for example, preprogrammed mechanical ‘breathing’ routines that users follow~\cite{matheus2022social}, huggable robots that sense respiration and aim to slow it to a calmer rhythm~\cite{lotzin2025feasibility}, soft-robotic surfaces embedded in the environment that guide breathing~\cite{sabinson2024every}, and robots that gradually decelerate a matched rhythm to deepen respiration~\cite{taminami2020effects}. Even humanoid robots acting as breathing coaches have been linked to more positive affective responses than comparable non-robotic interfaces~\cite{klkeczek2024robots}.

\section{Design and Implementation}

Our prototype includes a robotic arm that can perform a "breathing" motion, slightly elevating and lowering in sync with a respiration signal. The signal is acquired using a respiration sensor customized to the specific implementation. The signal can either be streamed directly to the robotic arm for synchronized breathing with the user, or played from a recorded breathing file for non-synchronized breathing. The robotic arm can also be controlled simultaneously with a game controller, allowing the selection and individual control of each degree of freedom (DoF), and enabling the performance of simple operations.

A custom interaction control software links the different components of the system and allows the experimenter to switch smoothly between operation modes.

We chose respiration as a biosignal that could enhance embodiment as recent research had shown similar effects in avatars~\cite{goral2024enhancing,wald_breathing_2023,wald_enriched_2025}. A key rationale for the use of respiration is that it is more distinctive in the capacity of the user to control it, in particular when compared with other signals tied to the autonomous nervous systems such as skin conductance or heart rate. Therefore, it may be more associated with agency, which in turn could potentially enhance the sense of embodiment. The representation of the respiration signal in movement, rather than other means such as light or sound, was done due to its embodied nature, and the powerful effects of robotic movement, whether biomimetic or choreographed, in HRI~\cite{Hoffman2015Design, bewley_designing_2018}. We designed the movement so that it resembles human respiration, going up and down with inhalation and exhalation, in order to enable an intuitive interpretation of the movement as breath.

\subsection{Technical implementation}

Our system is comprised of an Arduino Tinkerkit Braccio robotic arm, a modified BreathCli respiration sensorp~\cite{wald_enriched_2025}, a PlayStation 5 game controller, and a laptop running our interaction control software.  

The \textit{robotic arm} chosen was an Arduino Tinkerkit Braccio. It is controlled via an Arduino UNO microcontroller, programmed with an updated library BraccioV2~\footnote{https://github.com/kk6axq/BraccioV2} that provides extended functions for manipulating the movements of its different DoF (Base, Shoulder, Elbow, Wrist, Wrist rotation, and Gripper). Each joint of the robotic arm has a specific mechanical range of motion. Commands are transmitted via a serial port as motor angles. The effect of respiration was achieved using the movement of only the shoulder and elbow, moving in opposite directions so as to balance their overall movement and minimize its functional effect. 

The \textit{respiration sensor} chosen was the BreathClip. An open source IMU-based sensor that is either clipped or attached with a magnet on the user's shirt, over the collarbone. Designed for interaction~\cite{wald_enriched_2025}, it is based on the widely available M5StickC PLUS2 ESP32 Mini IoT Development Kit. It was chosen for its convenience of use with participants, ease of integration with the whole system, and for the ability to manipulate its firmware. 

The original sensor implementation truncates signals during acquisition, cutting off and reconstructing the signal in order to clearly mark stages while keeping the signal amplitude within a predefined range. For our implementation we edited the firmware to record the complete periodic fluctuations of the breathing waveform. 
This was done to ensure the inter-individual variability in respiratory amplitude is preserved. 

We also chose to use only a first-order filter rather than (an also available) second-order filter, opting for lower computational cost and latency on account of better noise reduction. This was done considering our scenario that requires a high level of synchronization while performing in a stationary, low-noise setting.

The \textit{Interaction control software}, integrates the different system elements, linking the respiration sensor, game controller, and robotic arm. The software architecture was organized around three components: an input terminal, a host, and an output terminal.

The input terminal collected signals from two sources: respiration data transmitted over Wi-Fi from the breathing sensor and manual control inputs received via USB from the game controller. These streams were buffered and forwarded to the host for further processing.

The host acted as the central processing unit, responsible for integrating signals, managing experimental conditions (synchronized vs. non-synchronized breathing), and providing the experimenter interface. It also maintained a shared state of the joint positions of the robotic arm. Communication between modules used lightweight protocols: UDP packets carried respiration data, USB provided controller inputs, and serial communication linked the host to the robotic arm.

The output terminal formatted the host’s commands into a six-element array, each representing one joint of the robotic arm (base, shoulder, elbow, wrist, wrist rotation, and gripper). At a fixed update rate, the array was serialized and sent over the serial port to the Arduino, which parsed the string into joint angles and drove the actuators accordingly.

In practice, these three conceptual modules were implemented as independent threads within the host architecture. Five threads managed the system concurrently: (1) reception of respiration data, (2) signal processing, (3) data storage, (4) game controller input handling, and (5) generation of motor instructions for the robotic arm. This thread-level independence ensured non-blocking operation, allowed respiration signals and controller inputs to be integrated in real time, and enabled smooth switching between synchronized and non-synchronized breathing conditions during the experiment.

For enabling the non-synchronized condition, pre-recorded breathing signals were collected from two male and two female pretest participants in a resting state. For each experimental session, four 30 s continuous segments were randomly selected from this pool of recordings, with each segment drawn from a single participant's recording file. The selected segments were then concatenated in a random order to generate a 2-minute pre-recorded breathing signal file, which was looped throughout the non-synchronized condition.

Under both conditions, the robot can be operated using the controller. The horizontal movement of the left joystick controlled the base rotation. By pressing the X, Square, or Triangle buttons, participants could select the shoulder, elbow, or wrist joint, respectively, after which the vertical movement of the left joystick controlled the corresponding joint’s motion. The right joystick was used to operate the gripper: horizontal movement controlled rotation, while vertical movement controlled opening and closing.

\textit{Mapping between breathing signals to joint movement} was performed as follows: Signal post-processing is performed in a dedicated thread. A fixed-length sliding window of $N=10$ samples is employed, such that every 10 consecutive data are processed jointly. For a window segment ${x_{t}, x_{t+1}, ...,x_{t+N-1} }$, the integration is computed as 
\begin{equation}
I_k = \sum_{i=0}^{N-1} x_{t+i}
\end{equation}
where $I_k$ denotes the integration value of the k-th window. The integration difference is then defined as 
\begin{equation}
\Delta I_{k} = I_{k} - I_{k-1}
\end{equation} 
which captures the rate of change in respiratory activity between successive windows. From a signal-processing perspective, this highlights changes in the rate of respiratory fluctuations while reducing the influence of baseline shifts or slow drifts.
Physiologically, $\Delta I_{k}$ can be interpreted as capturing the acceleration of inhalation and exhalation, thereby providing a more direct correspondence between respiratory effort and the resulting joint movements of the robotic arm.
To obtain a standardized measure, the integration difference is normalized with respect to predefined minimum and maximum reference values $\Delta I_{min}$ and $\Delta I_{max}$: 
\begin{equation}
\Delta I^{*}_{k} = \frac{\Delta I_{k}-\Delta I_{min}}{\Delta I_{max}-\Delta I_{min}}\cdot 2 -1
\end{equation}
such that $\Delta I^{*}_{k}\in[-1,1]$. The normalized integration difference $\Delta I^{*}_{k}$ is subsequently mapped to the angular displacement of the robotic arm joints. For the shoulder joint, the angular displacement is given by
\begin{equation}
\theta_{shoulder}(k)=\Delta I^{*}_{k}\cdot\theta^{max}_{shoulder}
\end{equation}
where $\theta^{max}_{shoulder}=6^\circ$ denotes the maximum angular displacement per one movement. The sign of $\Delta I^{*}_{k}$ determines the direction of movement. For both joints, a negative angular displacement corresponds to a downward movement, whereas a positive angular displacement corresponds to an upward movement.  
The elbow joint is modeled with its displacement proportional to the normalized shoulder displacement: 
\begin{equation}
\theta_{elbow}(k)=\Delta I^{*}_{k}\cdot\theta^{max}_{elbow}
\end{equation} where $\theta^{max}_{elbow}=4^\circ$
. To achieve smoother and more natural kinematic behavior, the maximum angular displacement of the elbow joint is deliberately restricted to a smaller range than that of the shoulder joint. The other joints of the robotic arm are not coupled to respiratory signals and therefore remain unaffected. 
The calculated angular displacement is directly mapped onto the corresponding joint of the robotic arm at its current position. 
This displacement is then incrementally added to the joint’s existing angle to determine the updated absolute position of each joint at the current time step.

\section{Methodology}

\subsection{Participants}
The research involved $n = 26$ participants, comprising $11$ females, $14$ males, and $1$ identifying as other, with an average age of $30.73 \pm 6.81$. Participants were recruited from the university student pool by convenience sampling and received a snack for their participation. Before starting the experiment, informed consent was obtained from each participant. All individuals reported normal vision, hearing, and neurological function, alongside fluency in English.

\begin{figure}[tb]
  \centering
  \includegraphics[width=\linewidth]{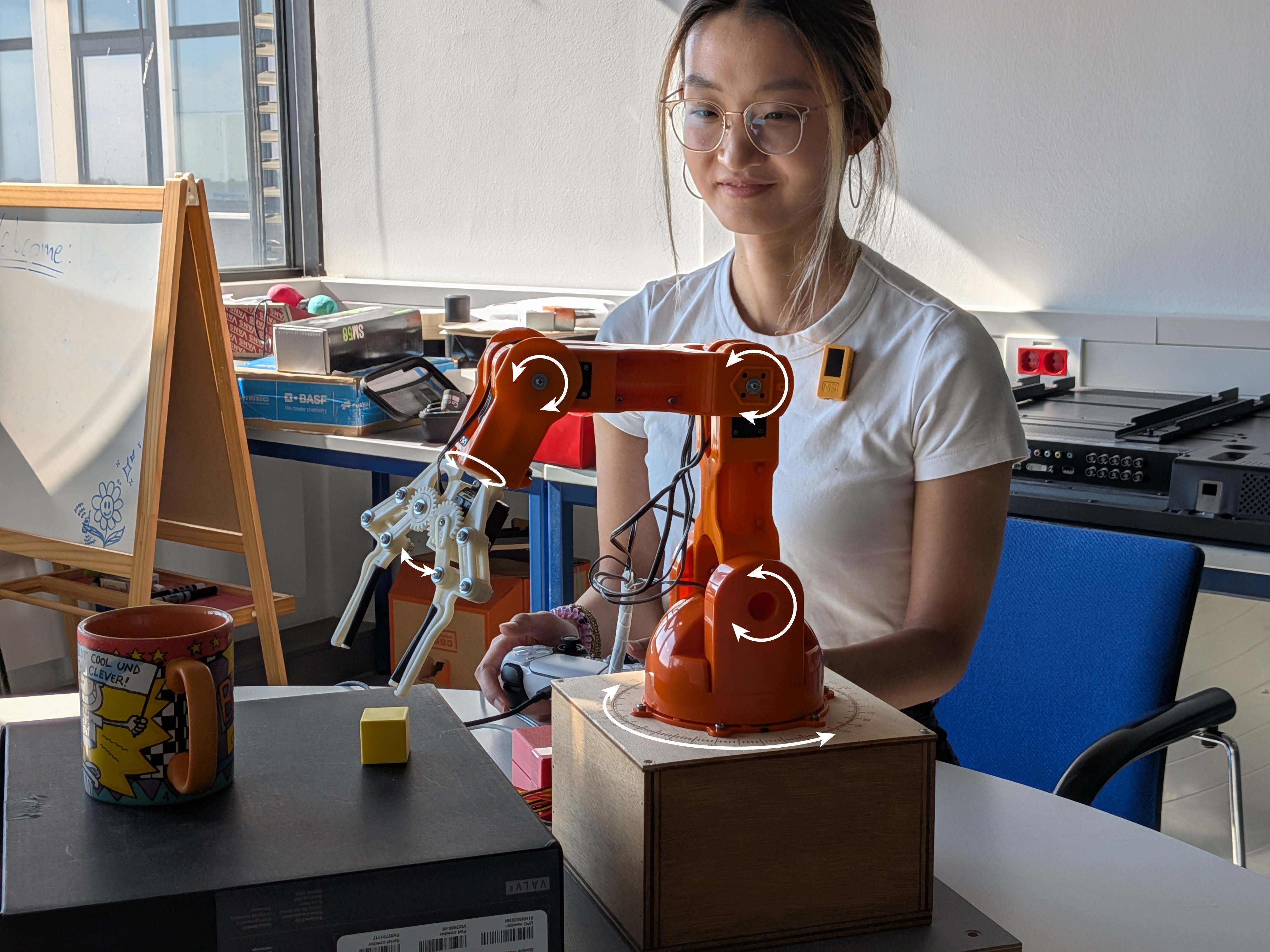}
  \caption{The experiment setup, during the introduction phase. Participants were asked to grasp the yellow cube and place it in the cup. The DoFs are indicated by the overlaid white arrow. From base to gripper: Base, Shoulder, Elbow, Wrist, Wrist rotation, and Gripper.}
  \label{fig:Setup}
  \Description{A woman sits at a table operating an orange robotic arm during the experiment's introduction phase. She wears a respiration sensor clipped to her shirt and holds a controller in her hand. In front of the robot are a colorful mug and a small yellow cube. Overlaid white arrows on the robot indicate its five degrees of freedom (DoFs), labeled from base to gripper: Base rotation, Shoulder, Elbow, Wrist, Wrist rotation and Gripper. The participant’s task was to grasp the yellow cube and place it in the mug.}
\end{figure}

\subsection{Materials}

Participants’ subjective experiences were evaluated through a battery of standardized questionnaires, including:

\begin{itemize}
    \item \textbf{Virtual Embodiment Questionnaire (VEQ):} The questions were ranked on a 1-7 Likert scale\cite{roth2020construction}. The questions were adapted to the phrasing "robotic arm" instead of "virtual body". 
    \item \textbf{Rubber Hand Illusion Questions (RHI):} We employed questions adapted from pre-existing questionnaires utilized in rubber hand illusion experiments~\cite{kalckert_moving_2012}. The questions were ranked on a 1-7 scale. The questions were adapted to the phrasing "robotic arm" instead of "rubber hand".  
    \item \textbf{Self-Efficacy in HRI Scale:} We employed the short version of the Self-Efficacy in HRI Scale~\cite{putten2018development}, rated as a scale from 0-100.  
    \item \textbf{Godspeed subscales:} We employed the anthropomorphism, animacy, and perceived safety subscales from the Godspeed questionnaire in their entirety, and used single additional items from the likeability scale~\cite{bartneck2023godspeed}. The questions were ranked on a 1-5 scale.

\end{itemize}

In addition, \textbf{open-ended questions} elicited qualitative reflections on participants’ experiences, providing complementary insights beyond the standardized instruments.

\begin{figure*}[t]
  \centering
  \includegraphics[width=\linewidth]{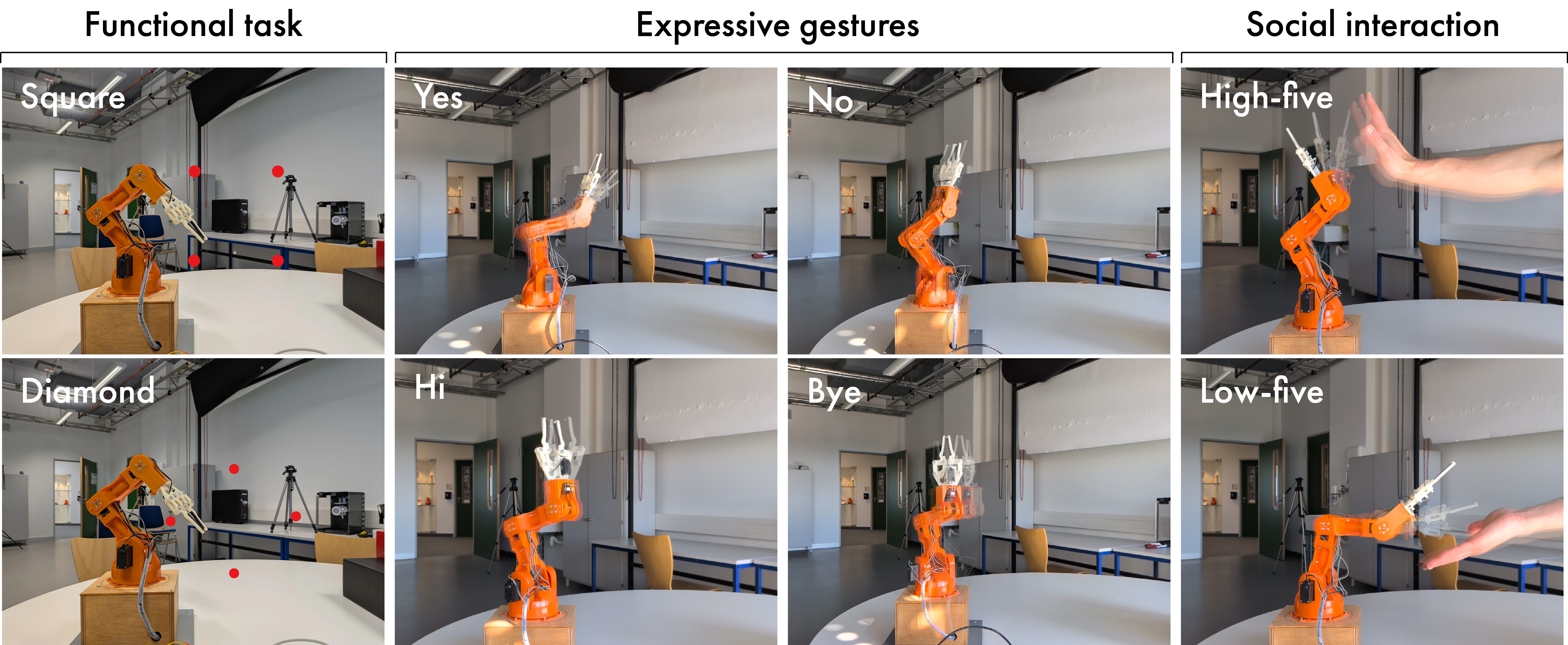}
  \caption{Illustrations of the tasks participants performed during the experiment's block task phase. With the order randomized between conditions per-participants, each each condition included a functional task, two expressive gestures, and a social interaction: draw the shapes square/diamond shape in space;  gestures yes and no / gestures hi and bye; give a high-five/low-five. For the drawing task, participants were presented with the above images and were instructed to reach the four points.}
  \label{fig:Tasks}
  \Description{A 3x3 grid of images shows a robotic arm performing tasks grouped into three categories: Functional task, Expressive gestures, and Social interaction.

Functional task column: The top image is labeled "Square," showing the robotic arm drawing an invisible square in space, with four red dots marking corner points. The bottom image is labeled "Diamond," showing the robot tracing a diamond shape with similar red markers.

Expressive gestures column: The top image is labeled "Yes," with the robot's arm posed in a vertical nodding motion. The middle image, labeled "No," shows a side-to-side gesture. The bottom image, "Hi," shows the arm raised with open fingers, and "Bye" shows a waving motion through overlapping hand positions.

Social interaction column: The top image is labeled "High-five," where the robot's hand meets an approaching human hand in midair. The bottom image is labeled "Low-five," where the human hand is positioned palm-up, and the robot's hand moves downward to meet it.

These images illustrate tasks participants imitated during the experiment, with conditions randomized per participant.}
\end{figure*}

\subsection{Study design}

To assess the impact of real-time breathing synchronization on embodiment, we employed a within-subjects design with two counterbalanced conditions: synchronized breath, and non-synchronized breath. In both conditions, participants operated the robotic arm using the handheld controller to complete specified tasks. In the synchronized condition, the robotic arm's movement was synchronized with the participant's own breathing. 

Participants first signed an informed consent form and listened to an oral explanation of the research purpose and tasks. After signing, the experimenter demonstrated how to attach the sensor to the participant's shirt collarbone area with a magnet. It was also explained that the robotic arm's movements may or may not respond to their breathing.

Participants were introduced to the robot, and the use of the controller was explained to them. They were then instructed to freely move the robotic arm to ensure they understood the operation interface. Once they indicated their familiarity with the controls, they were asked to perform a simple practice task: grasping a 3D-printed yellow cube and placing it in a cup positioned behind the cube (see ~\autoref{fig:Setup}). The task was performed with no breathing signal presented, and was meant to ensure a similar basic performance proficiency level for all participants in controlling the robotic arm. After the participants successfully completed the introductory phase, the experimenter told the participant that the experimental session is about to begin and activated the breathing sensor.

Each experimental condition consisted of three phases:
\begin{enumerate}
    \item \textbf{Acclimatization Period (1 minute):} Participants quietly observed the robotic arm “breathing” for one minute without interacting with it. The mode depended on the pre-assigned conditions, synchronized or non-synchronized.  

    \item \textbf{Block Task Phase (approximately 3-5 minutes):} Participants were asked to perform a series of four tasks (see~\autoref{fig:Tasks}). We intentionally chose tasks with no clear success or failure criteria, in order to minimize the effect of objective performance on the participant's perceived performance and sense of control. The tasks included a functional task, two expressive gestures, and a social interaction, each randomized for each participant: draw a square/diamond shape in space; gestures yes and no/hi and bye; give a high-five/low-five.

    \item \textbf{Questionnaire Phase:} Following each condition, the breathing representation was paused, and participants completed the questionnaires.
\end{enumerate}

After completing both conditions, participants answered comparison questions regarding their preferences and perceived differences between conditions. Between sessions, the robotic arm and controller were reset and sensors recharged for the next participant.

\begin{figure}[b]
  \centering

  \begin{subfigure}{0.49\linewidth}
    \centering
    \includegraphics[width=\linewidth,trim=0 0 0 35,clip]{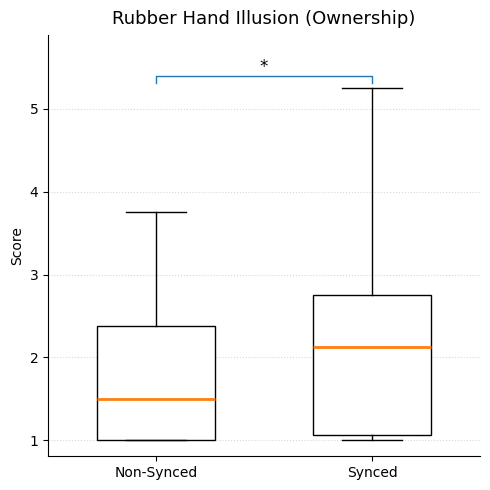}
    \caption{RHI Ownership}
    \label{fig:rhi}
  \end{subfigure}
  \hfill
  \begin{subfigure}{0.49\linewidth}
    \centering
    \includegraphics[width=\linewidth,trim=0 0 0 35,clip]{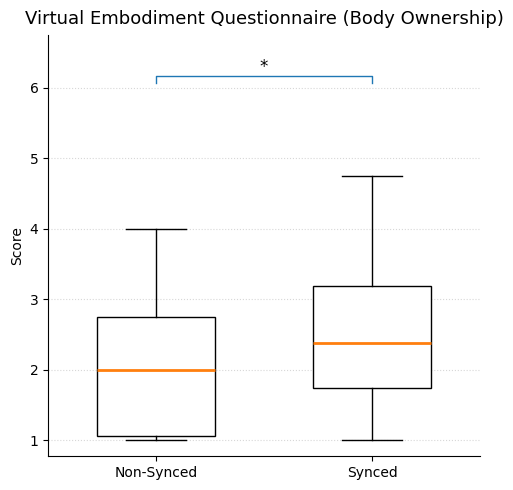}
    \caption{VEQ Body Ownership}
    \label{fig:veq}
  \end{subfigure}

  \caption{Boxplots presenting the significant scores across conditions (non-synchronized vs. synchronized).}
  \Description{Two side-by-side boxplots compare participant scores between non-synchronized and synchronized conditions.

(a) RHI Ownership: The boxplot shows higher median and overall scores in the synchronized condition compared to the non-synchronized one.

(b) VEQ Body Ownership: Similarly, the synchronized condition shows higher scores than the non-synchronized condition.
Both plots display asterisks above the comparisons, indicating statistically significant differences.}
  \label{fig:boxplots}
\end{figure}


\section{Results}

\subsection{Quantitative Analysis}

We compared participants’ responses across the two experimental conditions (Non-Synchronized vs Synchronized) using one-sided paired-samples $t$-tests (alternative = greater) due to our directional hypothesis. Shapiro--Wilk tests indicated that most difference scores did not significantly deviate from normality, supporting the use of paired $t$-tests. To control for multiple comparisons across subscales, $p$-values were adjusted using Holm--Bonferroni correction ($\alpha = .05$). Cohen’s $d$ is reported as a standardized effect size. To check for potential order-effects, we tested whether results differed depending on whether participants first experienced the Non-Synced or the Synced condition. 

Two measures of ownership survived Holm--Bonferroni correction. Participants reported a stronger sense of ownership in the VEQ -- Body Ownership subscale (Non-Synced = 2.12, Synced = 2.59, $\Delta = 0.47$, $t(25) = 3.01$, $p = .0029$, $p_{\mathrm{holm}} = .0265$, $d = 0.59$) and in the RHI -- Ownership subscale (Non-Synced = 1.80, Synced = 2.16, $\Delta = 0.36$, $t(25) = 3.05$, $p = .0027$, $p_{\mathrm{holm}} = .0265$, $d = 0.60$). Both effects indicate a medium-to-large increase in ownership in the Synced condition relative to the Non-Synced condition.

All other measures, including perceived safety, self-efficacy, animacy, agency, anthropomorphism, did not show significant differences after correction. Order-effect checks confirmed that starting condition did not alter the pattern of results. These findings indicate that the observed differences reflect genuine condition effects rather than artifacts of presentation order.

\subsection{Qualitative Analysis}

We conducted an inductive thematic analysis of the open-text responses following accepted guidelines~\cite{braun_using_2006}. In the first stage, participants’ comments were coded to capture recurring ideas and experiences using Atlas.ti. These initial codes were then organized according to the experimental condition, allowing us to compare responses made in the Synced, Non-Synced conditions, and general comments made about the interaction with the robot. In the final stage, related codes were clustered into higher-level themes that captured broader patterns across the data, resulting in five overarching categories: agency and control, discomfort and frustration, awareness of breathing, and social framing.

Most participants indicated that they preferred the synchronized condition (14/26 - 54\%), compared to fewer who favored the Non-synchronized condition (9 - 35\%), while one participant indicated liking both equally, one preferred neither, and one left the question blank. The majority also reported noticing the responsiveness of the robotic arm (20 - 77\%), suggesting that synchrony or lack of synchrony were salient features of the interaction. 

Participants could explicitly describe when the robot mirrored their breathing, with statements like: ``understood it’s with my breathing'' [P2], ``breathing as I breathed'' [P16], and ``responded to my breathing'' [P17]. Awareness was mentioned less in regards to the  non synced conditions, describing it as it ``wasn’t my breathing'' [P28], or that the arm had ``its own breathing'' [P17].

The theme of agency and control was identified by many participants indicating the saliency of the experience of controlling the robot. For both conditions agency was mentioned in a positive sense with a few negative comments, thought the arguments for each were different.  In the Synced condition, agency mentions related to its providing an ``additional control somehow'' [P28] and viewing it as ``another tool'' [P5], only two participants describing the movement as ``unexpected'' [P26] or ``difficult'' [P6]. In the non-synced condition, participants likewise described the robot as easy to manage and predictable, ``easier to handle'' [P27], ``easier because it was always the same'' [P28]. At the same time, in this condition there was mention of the robot’s independence, remarking that it ``had its own mind'' [P17]. 

There were slightly more mentions of discomfort in the non-synced condition than in the synced condition. Participants described similar mismatchs and difficulties for both conditions, using phrases such as “did not match” [P2], "more rigid" [P16, P2] “doesn't quite do what you want” [P5], “random” [P8, P23], and “annoying” [P23]. 

Participants also described the experience in social or expressive terms. Most of these were general remarks not tied to a condition, but when they appeared with respect to a condition, they appeared more often in the synchronized condition. They described the interaction with the robot in the synced condition as “It felt nicer” [P16], “more interesting” [P17]. 

\section{Discussion}

Our results show that synchronizing a robot’s movements with a user’s breathing reliably enhances the sense of embodiment. Participants reported stronger feelings of body ownership in the synchronized condition, supported by medium-to-large effect sizes across two standardized measures for ownership. Qualitative accounts reinforced these findings, with the majority of participants indicating that they preferred the synced condition. 

Taken together, our findings provide clear support for RQ1: real-time breath synchronization enhanced the sense of embodiment in teleoperation, significantly increasing ownership of the robotic arm. This result demonstrates that interoceptive-exteroceptive synchrony can serve as a powerful new mechanism for fostering a sense of ownership and embodiment in human–robot interaction. With respect to RQ2, we did not find significant evidence that breathing synchrony influenced self-efficacy or social impressions, suggesting that these broader effects may depend on more complex or sustained contexts. Nevertheless, social descriptions were used in general and for the synced condition, yet hardly at all for the non-synced condition.

Our results open a promising avenue for prosthetic technologies, where embodiment is directly tied to daily use and psychosocial outcomes~\cite{graczyk2018home}. Despite the human body’s remarkable capacity to embody artificial limbs, less than half of amputees use their prostheses daily~\cite{jang_survey_2011}. Challenges remain at multiple levels: phenomenological (a subjective sense of the prosthesis as one’s own), cognitive (adoption of the prosthesis into one’s body schema), and neural (capitalizing on brain plasticity)~\cite{makin_neurocognitive_2017, segil2022measuring}. Most existing noninvasive feedback approaches, such as vibrotactile, auditory, or visual cues time-locked to prosthetic events, are designed to improve functional performance, while their impact on embodiment is often secondary and only rarely assessed directly~\cite{zbinden2022prosthetic, segil2022measuring}. Our results point to interoceptive-exteroceptive synchrony as a complementary pathway to address this gap. Future work should investigate how integrating respiratory or other physiological signals into prosthetic control might foster stronger embodiment, thereby supporting long-term acceptance and daily use.

Beyond conventional prostheses, research highlights that embodiment can extend to devices with varying levels of autonomy. For example, Weinberg’s robotic prosthetic arm for drumming integrates EMG-based control with semi-autonomous behavior, allowing one stick to follow the user’s muscle signals while another improvises in response to the music~\cite{weinberg2020robotic}. Such designs illustrate how prostheses can act not only as replacements but as creative extensions of the self, blending human and robotic agency. This perspective underscores that embodiment is not confined to teleoperation: it can also be cultivated in assistive and collaborative robots, where interoceptive-exteroceptive synchrony offers new opportunities to support integration between human and machine. Use cases include the integration of interoceptive-exteroceptive synchrony in semi-autonomous prostheses or assistive robots integrated into wheelchairs (e.g Kinova assitive robotics and EDEN~\cite{Hochberg2012Reach,Vogel2015assistive}).

User embodiment can also be central in collaborative human–robot interaction. In collaboration, people come to experience agency and ownership toward robotic partners when they perceive the robot’s intentions as aligned with their own and when sensorimotor contingencies reliably link their actions to the robot’s responses~\cite{roselli2022human, navare2024performing, sato2018body}. Building on this line of work, our study examined interoceptive synchrony. Specifically aligning breathing rhythms, as a novel pathway to strengthen the sense of shared embodiment. Such perceived embodiment allows users to experience robot actions as not only coordinated with their own but also partly belonging to them, which in turn fosters trust and mutual awareness in collaboration~\cite{cross2019social}. Further possible directions for exploration are in collaboration with other humans and the robots they feel a sense of embodiment towards, as well as exploring the notion of embodiment in multiple robots. 

Participants also described the interaction experience in social terms, moreso specifically in the synced condition. This suggests that respiration synchrony may foster a stronger sense of interaction and connection. Such socially oriented attributions align with the CASA framework \cite{nass1994computers}, which proposes that people often respond to computers and robots as if they were social partners, especially when the system displays responsive cues, in this case respiration synchrony. This observation is further supported by the finding that even minimal robotic movement and gestures are automatically interpreted as social cues \cite{erel_robots_2019} even in non-humanoid robots \cite{anderson-bashan_greeting_2018, zuckerman_companionship_2020}.

Understanding embodiment in collaboration also sets the stage for shared autonomy, where the balance of human and robotic control determines also how much agency the human partner feels. Shared autonomy is central in human–robot collaboration because tasks often require balancing human input with robotic assistance. For example shared autonomy and assistive robotics - where the robot blends the user’s commands with its own assistive control to complete a task - assistive robotic arms range from wheelchair-mounted manipulators used daily (e.g., JACO, iARM) to table-side/mobile arms for feeding and manipulation (PR2, Panda). While formal models describe shared autonomy as blending human and robot policies~\cite{dragan2013policy, javdani2018shared}, research increasingly shows that this balance is not only technical but experiential. Higher levels of robotic assistance can improve task performance but often reduce the user’s sense of agency~\cite{collier2025sense}. Phenomenological accounts emphasize that how people experience sharing control depends on the context of use and how the interaction unfolds over time~\cite{coeckelbergh2011humans}. This points to a core challenge: sustaining collaboration that is both effective and subjectively meaningful. We suggest that embodiment offers a way to bridge this trade-off. When robotic actions are experienced as part of one’s own, assistance can enhance rather than diminish agency. Interoceptive-exteroceptive synchrony could potentially enable increased synchrony with a robot even when not directly controlling its actions, and by so fostering this sense of shared embodiment, providing an approach toward fluent collaboration that preserves both performance and agency.

Our findings highlight the potential of interoceptive signal synchrony for strengthening ownership and sense of embodiment. While our work focused solely on breathing, future research should consider other interoceptive signals such as heartbeat or skin conductance, that could provide similar effects and would be more fitting for some use cases. Comparative studies are needed to determine which physiological pathways most effectively foster ownership and agency. Finally, our embodiment measures relied primarily on subjective reports, which may not capture the full range of interoceptive effects; future work could incorporate neural, physiological, and behavioral indicators to provide a more comprehensive assessment.

\section{Conclusion}

This work introduces embreathment as a new pathway for fostering embodiment in human–robot interaction. By synchronizing robot movements with user breathing, we demonstrated that breath synchronization can significantly strengthen body ownership during teleoperation. While broader measures of self-efficacy and social perception were not significantly affected, qualitative accounts suggested that synchrony shaped participants’ sense of control, comfort, and connection with the robot.

These findings highlight that embodiment is not only a function of visuomotor congruence, but can also be supported through subtle physiological cues. Exploring interoceptive-exteroceptive synchrony opens possibilities for prosthetics, collaborative robots, and shared autonomy, where agency and ownership are critical to meaningful use. Further research could employ additional physiological signals, and examine how these mechanisms play out in long-term, real-world contexts.

By grounding robotic interaction in shared bodily rhythms, embreathment points toward a design space where robots are not merely tools but extensions of the self.

\begin{acks}
To Robert, for the bagels and explaining CMYK and color spaces.
\end{acks}

\bibliographystyle{ACM-Reference-Format}
\bibliography{main}


\end{document}